%% file: root.tex
\documentclass[letterpaper, 10pt, conference]{ieeeconf}  % Comment this line out if you need a4paper

\IEEEoverridecommandlockouts                              % This command is only needed if 
\usepackage[table]{xcolor}
\usepackage{lipsum}
\usepackage{amsmath}
\usepackage{graphicx}
\usepackage{dblfloatfix}
\usepackage{booktabs}
\usepackage{multirow}
\usepackage{lscape}
\usepackage{amssymb}
\usepackage{pifont}
\usepackage{hyperref}
\usepackage[owncaptions]{vhistory}
\usepackage[ddmmyy]{datetime}
\usepackage{balance}

\let\labelindent\relax
\usepackage{enumitem}
\newlist{myenum}{description}{10}
\setlist[myenum]{labelindent=\parindent, leftmargin=*, label=(\roman*), align=left}
\setlist[myenum]{leftmargin=8pt}

\newlist{myenum2}{description}{10}
\setlist[myenum2]{labelindent=\parindent, leftmargin=*, label=(\roman*), align=left}
\setlist[myenum2]{leftmargin=0pt}

\usepackage[doi=false,isbn=false,url=false,date=year,backend=biber,style=ieee,sorting=none,maxnames=3,minnames=1]{biblatex} 
\AtBeginBibliography{\small}
\AtEveryCitekey{\clearlist{publisher}}
\AtEveryBibitem{\clearlist{publisher}}

\AtEveryBibitem{%
  \ifentrytype{inproceedings}{%
    \clearfield{pages}%
  }{%
  }%
}

\DeclareSourcemap{ 
    \maps[datatype=bibtex]{
      \map{
      \step[fieldsource = booktitle,
          match = Computer Vision and Pattern Recognition,
          replace = CVPR]
      \step[fieldsource = booktitle,
          match = International Conference on Computer Vision,
          replace = ICCV]
      \step[fieldsource = journal,
          match = Intelligent Transportation Systems,
          replace = ITS]
        }
    }      
}

\title{\LARGE \bf
Scenario-based Evaluation of Prediction Models for Automated Vehicles
} 
\author{Manuel Muñoz Sánchez$^{1}$ Jos Elfring$^{2}$ Emilia Silvas$^{3}$ and René van de Molengraft$^{1}$% <-this % stops a space
\thanks{This work was supported by SAFE-UP under EU's Horizon 2020 research and innovation programme, grant agreement 861570.}% <-this % stops a space
\thanks{$^{1}$Manuel Muñoz Sánchez, Emilia Silvas, Jos Elfring and René van de Molengraft are with the Department of Mechanical Engineering, Eindhoven University of Technology, Eindhoven, The Netherlands.}%
\thanks{$^{2}$Jos Elfring is also with the Product Unit Autonomous Driving, TomTom, Amsterdam, The Netherlands.}%
\thanks{$^{3}$Emilia Silvas is also with the Department of Integrated Vehicle Safety, TNO, Helmond, The Netherlands.}%
}

\begin{document}

%%%%%% VERSION HISTORY
% \onecolumn
% \vspace*{7cm}
% \begin{versionhistory}
% \vhEntry{0.1}{23/02/22}{}{First draft with main document structure.}
% \vhEntry{0.2}{28/02/22}{}{First round of feedback and filled most incomplete sections.}
% \vhEntry{0.3}{01/03/22}{}{Implemented (most) feedback. Completed Section II (Preliminaries). Formalized error computation. Added abstract. Overall updates for better consistency throughout the document.}
% % \vhEntry{0.X}{\today}{}{}
% \end{versionhistory}

% The latest version of the document can always be found at {\color{blue}\href{https://www.overleaf.com/project/620fafb1e903f349b274c4f2}{https://www.overleaf.com/project/620fafb1e903f349b274c4f2}}. If you start reviewing the document a while after it was sent out, please re-download it to get the latest version. Feel free to suggest edits and add your comments in the overleaf project directly (tracking changes is turned on). 

% % \vspace*{\fill}
% \setcounter{table}{0}
% \newpage
% \twocolumn
%%%%%%%%%%%%%%%%%%%%

\maketitle
\thispagestyle{empty}
\pagestyle{empty}

%%%%%%%%%%%%%%%%%%%%%%%%%%%%%%%%%%%%%%%%%%%%%%%%%%%%%%%%%%%%%%%%%%%%%%%%%%%%%%%%
\begin{abstract}
To operate safely, an automated vehicle (AV) must anticipate how the environment around it will evolve. For that purpose, it is important to know which prediction models are most appropriate for every situation. 
Currently, assessment of prediction models is often performed over a set of trajectories without distinction of the type of movement they capture, 
resulting in the inability to determine the suitability of each model for different situations. 
In this work we illustrate how standardized evaluation methods result in wrong conclusions regarding a model’s predictive capabilities, preventing a clear assessment of prediction models and potentially leading to dangerous on-road situations. We argue that following evaluation practices in safety assessment for AVs, assessment of prediction models should be performed in a scenario-based fashion. 
% To provide a first step towards a scenario-based assessment of prediction models [Text updated]
To encourage scenario-based assessment of prediction models
and illustrate the dangers of improper assessment, we categorize trajectories of the Waymo Open Motion dataset according to  the  type  of  movement  they  capture. Next, three different models are thoroughly evaluated for different trajectory types and prediction horizons.  
Results show that common evaluation methods are insufficient and the assessment should be performed depending on the application in which the model will operate.

\end{abstract}

%%%%%%%%%%%%%%%%%%%%%%%%%%%%%%%%%%%%%%%%%%%%%%%%%%%%%%%%%%%%%%%%%%%%%%%%%%%%%%%%

\section{Introduction}
Automated vehicles (AVs) have become popular in recent years since they have the potential to increase road safety, efficiency and comfort \cite{Morando2018StudyingMeasures,Ploeg2011ConnectReduction,Milakis2017PolicyResearch}. To operate safely, an AV must accurately anticipate the future motion of other road users (RUs) in its surroundings~\cite{Rasouli2020DeepSurvey}.
To build trajectory prediction models, deep learning (DL) techniques are gaining attention \cite{Rudenko2020HumanSurvey}, since they can effectively learn complex interactions between different RUs \cite{Alahi2016SocialSpacesCustom,Gupta2018SocialNetworksCustom} and the road infrastructure
% Chandra2019Traphic:InteractionsCustom
\cite{Konev2021MotionCNN:Driving,gu2021densetntCustom} from past observations to produce more accurate predictions. Traditionally, training these models effectively was a problematic task since the amount of data required was not easily available. However, this issue has been alleviated in recent years with the release of several large public datasets \cite{Ettinger2021LargeDataset,Chang2019Argoverse:MapsCustom,Caesar2020NuScenes:DrivingCustom,kesten2019lyft,Zhan2019INTERACTIONMaps}. 
% These datasets are usually partitioned in different splits for different purposes. Some of the data is reserved for training the models (i.e. training split), some for evaluation of the models during training (i.e. validation split), and some for the final evaluation on data the models have not seen previously (i.e. test data). 
A common practice to assess a model's predictive accuracy is to consider a fraction of the dataset reserved for this purpose (commonly referred to as test data), and to compare the model's predictions with the real trajectories. 
% Various metrics exist to quantify the disparity between the real and predicted trajectories \cite{Rasouli2020DeepSurvey}, since models may vary significantly in the shape of their outputs. For example, some models produce a single prediction, while others produce a set of feasible trajectories and associated confidence for each. 
The output of prediction models may vary, hence different metrics exist to quantify the disparity between the real and predicted trajectories \cite{Rasouli2020DeepSurvey}. For example, some models produce a single prediction, while others produce a set of feasible trajectories and associated confidence for each. 

\begin{figure}[tb]
    \centering
    \includegraphics[width=\linewidth]{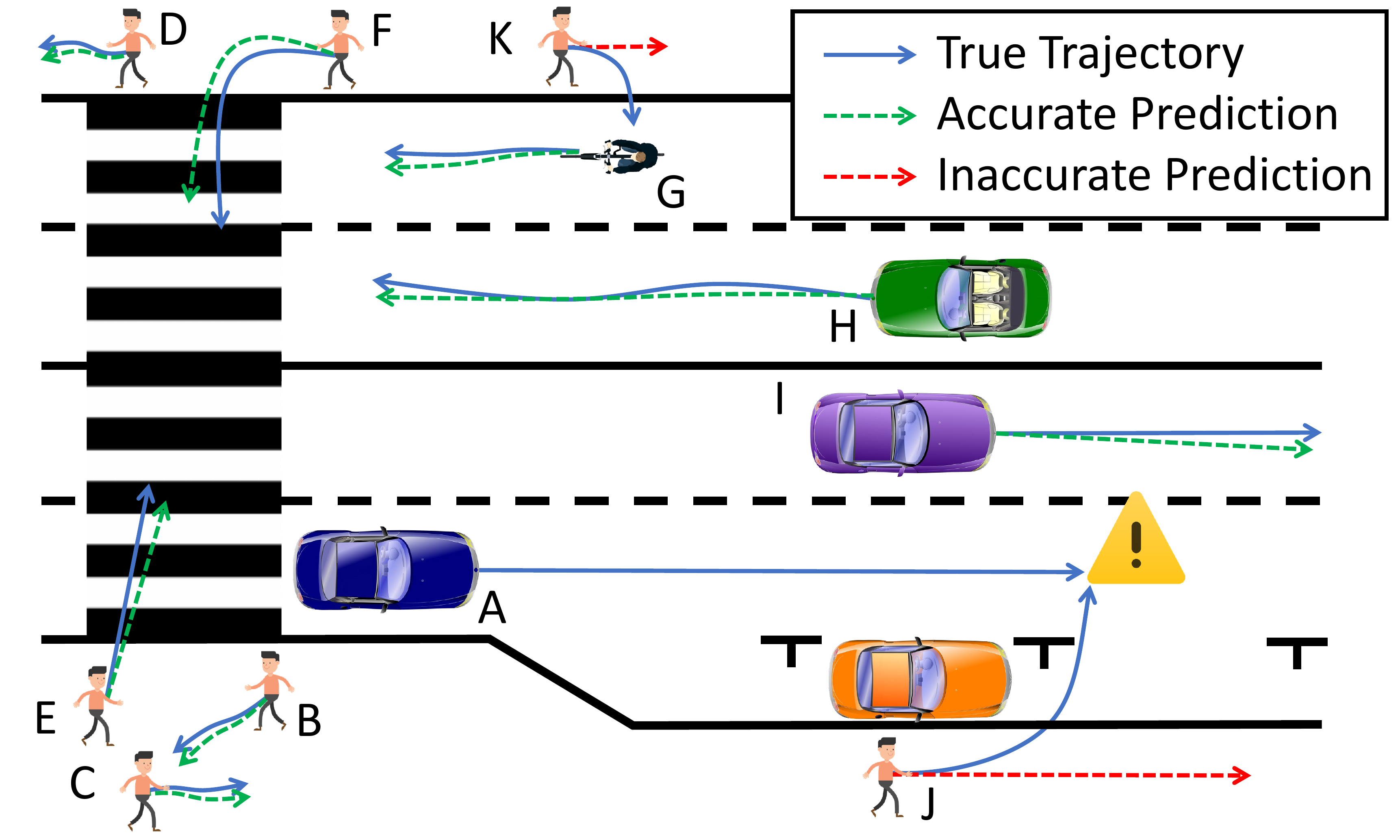}
    \vspace{-20pt}
    \caption{Example where a model that is accurate on average fails to predict a pedestrian trajectory, leading to a dangerous situation.}
    % \caption{Example of a crowded urban scenario where an AV (1) predicts the future trajectory of surrounding RUs (2-11). Current evaluation practices would deem this model suitable for RU trajectory prediction in crowded urban scenarios, since its predictions are highly accurate on average. It accurately predicts pedestrians on the sidewalk (2-4), crossing at designated crossings (5,6), and lane-following cyclists and vehicles (7-9). However, in this example only a few of these RUs are relevant to the AV (9, 10). Additionally, failure cases like the pedestrians crossing at non-designated crossings (10, 11) can go unnoticed since all trajectories are considered equally for error computation.}
    \label{fig:intro}
    \vspace{-20pt}
\end{figure}
\begin{figure*}[htb]
    \centering
    \includegraphics[width=0.9\linewidth]{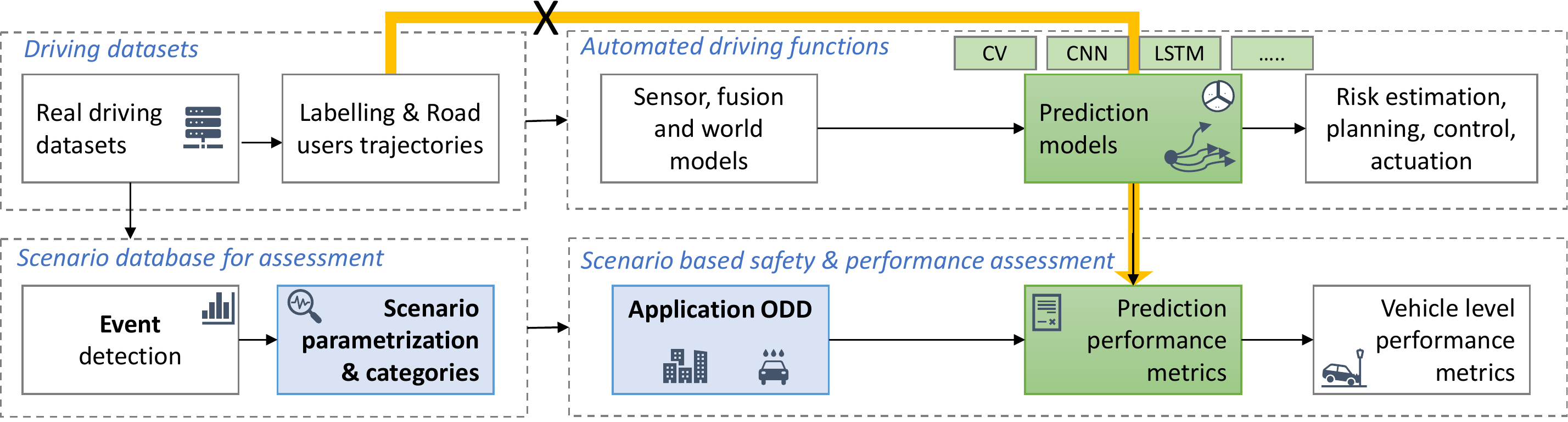}
    \vspace{-10pt}
    \caption{Overview of a scenario-based assessment pipeline. The orange arrow indicates the standard approach to evaluate prediction models. Green blocks concern prediction-specific activities. Blue blocks are currently lacking but required steps for thorough evaluation of prediction models. White blocks are relevant for vehicle level assessment, but out of the scope of this work.}
    \label{fig:overview}
    \vspace{-12pt}
\end{figure*}
Despite the existence of various evaluation metrics for prediction models, several challenges remain unaddressed in current evaluation practices, such as the inability of these metrics to capture a model's robustness or generalization capabilities \cite{Rudenko2020HumanSurvey}. Perhaps the most severe shortcoming is that all trajectories are considered equal for error computation despite capturing significantly different behaviors, which can lead to dangerours situations due to misjudgement of a model's suitability for specific situations. For instance, consider the situation shown in Fig.~\ref{fig:intro}, where an AV (A) predicts the future trajectory of surrounding RUs (B-K) in a crowded urban scenario. Current evaluation practices would deem this model suitable for RU trajectory prediction in crowded urban scenarios, since its predictions are highly accurate on average. It accurately predicts pedestrians on the sidewalk (B-D), crossing at designated crossings (E,F), and lane-following cyclists and vehicles (G-I). However, in this example only a few of these RUs are relevant to the AV (I, J). Additionally, failure cases like the pedestrians crossing at non-designated crossings (J, K) can go unnoticed since all trajectories are considered equally for error computation.

% (e.g. a pedestrian on the sidewalk vs. a pedestrian crossing at a non-designated crossing, as illustrated in Fig.~\ref{fig:intro}). 

The importance of a thorough evaluation for different types of trajectories has been recognized previously \cite{Kothari2021HumanPerspective}. However, current efforts to improve evaluation of prediction models
focus mainly on interactions between pedestrians (e.g. collision-avoidance \cite{Kothari2021HumanPerspective}), and disregard interactions of RUs with the road infrastructure (e.g. pedestrian stops at a red traffic light). Additionally, the evaluation procedure should provide a transparent assessment of a model's suitability for the intended application. For instance, for AVs, an inaccurate prediction for a pedestrian walking in front of the vehicle should be considered more important or severe than one of a pedestrian that is walking behind the vehicle or far from the road. 
As a second example, consider the purpose of vehicle fuel and energy optimization, where accurate long-term predictions are required for optimal path planning. On the contrary, for the development of emergency advanced driver-assistance system features (e.g. emergency braking or emergency steering) accurate short-term predictions become more relevant. Thus, it is important to assess the suitability of models with respect to the functional applications they will be used in, in other words, their operational design domain (ODD), and therein test for various scenarios and their overall impact with vehicle level performance metrics, not only prediction metrics (Fig. \ref{fig:overview}).

Current evaluation practices reporting averaged errors over all predicted trajectories are beneficial for ease of comparison between different models. If a model achieves a lower error over all the trajectories in the dataset, one can confidently say such a model is more accurate, at least on average. However, it remains unclear under which circumstances this model is preferred over others. To show that an improved assessment of prediction models is needed, and working towards that goal, the contribution of our work is as follows: 
\begin{enumerate}
    \item We illustrate the extent to which common evaluation methods, which only report average errors over all trajectories, result in misleading conclusions of a model's predictive capabilities, and argue that a scenario-based assessment is a more suitable approach. 
    \item We facilitate scenario-based evaluation of prediction models providing an open source framework\footnote{Code available at \scriptsize \nolinkurl{https://github.com/manolotis/SBEP}}, which will allow for a transparent evaluation of a model's capabilities for different situations, leading to an optimal choice of prediction model depending on the application.
    % [Text updated] \item We take a first step towards a scenario-based evaluation of prediction models\footnote{Code available at \scriptsize \nolinkurl{https://github.com/manolotis/SBEP}}, which will allow for a transparent evaluation of a model's capabilities for different situations, leading to an optimal choice of prediction model depending on the application.
\end{enumerate}

The remainder of this article is structured as follows. Section \ref{sec:preliminaries} introduces common trajectory prediction metrics and datasets, and presents related work on scenario-based evaluation. Section \ref{sec:method} introduces the prediction models compared and outlines how the comparison will be done using standard evaluation practices. Section \ref{sec:results} presents an analysis of the results, and Section \ref{sec:conclusion} concludes the work and highlights future improvements.

\section{Preliminaries} \label{sec:preliminaries}
This section summarizes the most commonly used performance metrics, recent datasets used to develop AV applications, and related work on scenario-based evaluation.

\subsection{Common Trajectory Prediction Metrics}
A plethora of performance indicators exist to evaluate trajectory prediction models~\cite{Rasouli2020DeepSurvey}, with average displacement error (ADE) and final displacement error (FDE) being the most popular~\cite{Rudenko2020HumanSurvey}. ADE measures the difference between the predicted and ground truth trajectories, averaged over all prediction horizons. FDE measures this difference at a specific horizon. To allow comparison of deterministic models that produce a single trajectory with probabilistic models that produce multiple feasible trajectories, variants of these metrics are used which report the errors of the trajectory that achieved the best accuracy. These variants are commonly referred to as minADE and minFDE. Although these metrics have several limitations~\cite{Rasouli2020DeepSurvey} and new metrics have been introduced recently to address some of these limitations~\cite{Ettinger2021LargeDataset}, we use them in this work since they remain the most common performance indicators at the moment. 

% To formally define these metrics, let N denote a set of road users, for which a prediction   for a set of timesteps T. 

To formally define these metrics, let $\hat{\mathbf{S}}$ denote a set of trajectory predictions for a set of road users $\mathbf{N}$ at future prediction horizons $\mathbf{T}$. The minADE of the predictions for prediction horizon $t$ is given by
\begin{equation}\label{eq:minADE}
    \text{minADE}(\hat{\mathbf{S}},t) =  \sum_{n\in \mathbf{N}} \min_{\hat{s} \in \hat{\mathbf{S}}^n} \sum_{\substack{t' \in \mathbf{T} \\ t' \leq t}} \frac{\Vert \hat{s}_t - s_t^n \Vert_2}{|\mathbf{N}| \times |\mathbf{T}|},
\end{equation}
where $\hat{\mathbf{S}}^n$ denotes the set of predictions for a road user $n$, $\hat{s}_t$ denotes the predicted position at time $t$, and $s^n_t$ denotes the true position of road user $n$ at time $t$. Additionally, $|.|$ denotes the size of a set and $\Vert . \Vert_2$ denotes the L2-norm of a vector. Similarly, the minFDE at a given prediction horizon $t$ is defined as
\begin{equation}\label{eq:minFDE}
    \text{minFDE}(\hat{\mathbf{S}}, t) =  \sum_{n\in \mathbf{N}} \min_{\hat{s} \in \hat{\mathbf{S}}^n} \frac{\Vert \hat{s}_t - s_t^n \Vert_2}{|\mathbf{N}|}.
\end{equation}

\subsection{Recent Datasets \& Waymo's Motion Prediction Challenge}
Several large datasets are publicly available for development and evaluation of prediction models, with some of the most recent and often used ones as summarized in Table~\ref{tab:dataset-overview}.
For this research we chose the Waymo Open Motion Dataset (WOMD) since it has the longest horizon, covers several cities, and contains extra information such as traffic light states. 

With the release of WOMD, the Waymo motion prediction challenge\footnote{\href{https://waymo.com/open/challenges/2021/motion-prediction/}{https://waymo.com/open/challenges/2021/motion-prediction/}} (WMPC) was introduced. In this challenge, the task is to predict the trajectories of a subset of RUs for 8 seconds into the future, given their history for the past 1 second and corresponding map of the area. The trajectories required to be predicted are selected to include interesting behavior and a balance of RUs as specified in \cite{Ettinger2021LargeDataset}.
In this work, we refer to those trajectories as \textit{trajectories to predict} (TTP), which make up about 7\% of all trajectories present in the dataset.
\begin{table}[htb]
\caption{Overview of datasets (adapted from \cite{Ettinger2021LargeDataset})}
\vspace{-6pt}
\label{tab:dataset-overview}
\resizebox{\linewidth}{!}{%
\begin{tabular}{@{}c|ccccc@{}}
\toprule
& Lyft  & NuSc  & Argo.  & Inter.  & WOMD  \\ \midrule
Reference article & \cite{kesten2019lyft} & \cite{Caesar2020NuScenes:DrivingCustom} & \cite{Chang2019Argoverse:MapsCustom} & \cite{Zhan2019INTERACTIONMaps} & \cite{Ettinger2021LargeDataset} \\
Prediction horizon {[}s{]} & 5    & 6        & 3         & 3            & 8     \\
Number of segments          & 170k & 1k       & 324k      & -            & 104k  \\
Segment duration {[}s{]}   & 25   & 20       & 5         & -            & 20    \\
Sampling rate {[}Hz{]}     & 10   & 2        & 10        & 10           & 10    \\
Cities                     & 1    & 2        & 2         & 6            & 6     \\
Map available              & \checkmark    & \checkmark        & \checkmark         & \checkmark            & \checkmark     \\
Traffic light states       & \checkmark    &          &           &              & \checkmark     \\ \bottomrule
\end{tabular}%
}
\end{table}
% \vspace{-12pt}

\subsection{Scenario-based Assessment}
The need for a more complete assessment of a model's predictive capabilities has been briefly recognized in previous works. Some authors recognize that the largest prediction errors occur in non-linear regions of the trajectory and report the ADE for these regions separately~\cite{Alahi2016SocialSpaces,Xu2018EncodingPrediction}. Some other works report on average maximum errors along the entire predicted trajectory to give an indication of worst-case predictions~\cite{Lee2017DESIRE:Agents,Felsen2018WhereAutoencoders}. However, these practices have not become the standard and lack the ability to capture relevant situations from the point of view of an AV. 

In the area of safety assessment for driver assistance systems and AVs, a scenario-based approach has been adopted. 
This approach presents several benefits, such as the ability to evaluate the coverage of the assessment, and the possibility of a direct translation between test outcomes and an assessment of the AV's performance with respect to a specific ODD, ultimately facilitating legal and public acceptance of AVs~\cite{DeGelder2022TowardsFrameworkCust}.

% , since it presents several benefits such as the ability to evaluate the coverage of the assessment~\cite{DeGelder2022TowardsFrameworkCust}, and allowing a direct translation of test outcomes into an assessment of the AV's performance with respect to a specific ODD~\cite{DeGelder2022TowardsFrameworkCust}, facilitating legal and public acceptance of AVs. 

In the area of trajectory prediction assessment, this scenario-based approach has not been generally adopted yet.
TrajNet++ is a recent benchmark with the goal of standardizing trajectory prediction applying similar concepts~\cite{Kothari2021HumanPerspective}. However, the focus is on interacting RUs (mainly pedestrians) and disregard other interactions with the road infrastructure (e.g. a traffic light). Additionally, specification of scenarios should consider relevant situations for the AV. For example, failing to predict the trajectory of a pedestrian that ends up on the road to avoid another pedestrian is irrelevant to the AV if this interaction occurs behind or far from the vehicle. However, if it occurs immediately in front of the vehicle, it would be highly relevant. The authors of \cite{Indaheng2021ASimulation} proposed a framework for scenario-based testing of prediction models in a simulated environment. The framework supports modeling and generation of scenarios involving interactive RUs for a thorough evaluation of prediction models. This approach overcomes several limitations of current evaluation methods. However, it is important to also evaluate how prediction models perform with real driving data, and the impact on the entire AV architecture with vehicle-level performance metrics.

\section{Methodology \& Experiments}\label{sec:method}
\begin{figure*}[htb]
    \centering
    \includegraphics[width=0.98\linewidth]{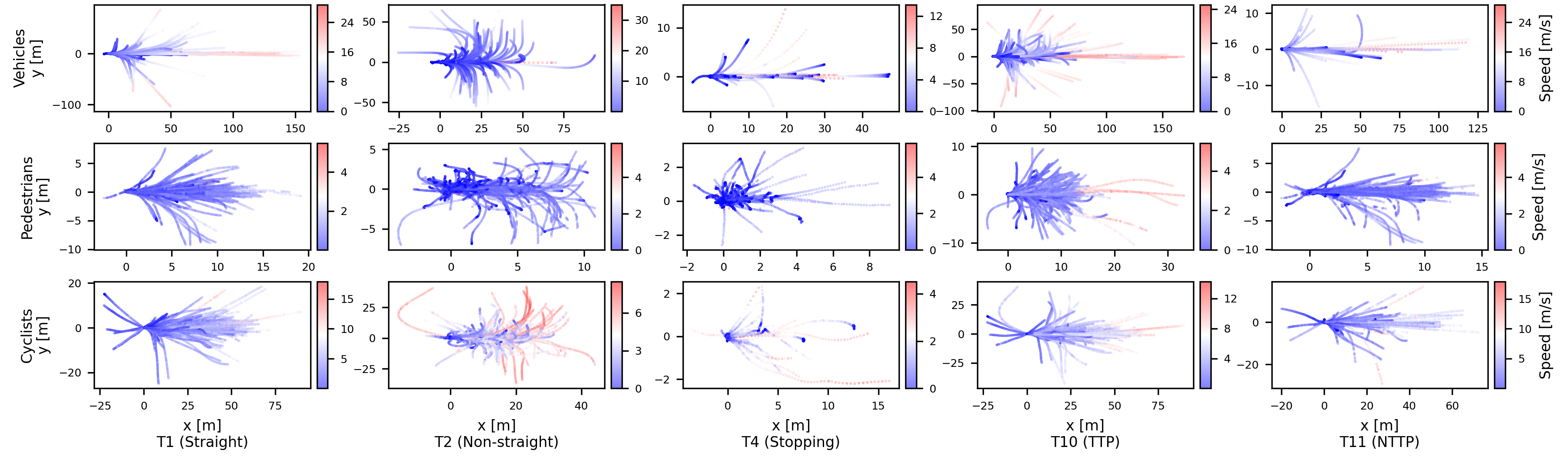}
    \vspace{-12pt}
    \caption{Examples of trajectories labeled with different tags. All RU trajectories start at the origin with heading pointing along the positive X axis.}
    \label{fig:trajectories}
    \vspace{-14pt}
\end{figure*}

To perform a thorough analysis of the performance of a prediction model for different types of trajectories, we first need to systematically detect these trajectory types in existing datasets. 
As a first step towards a scenario-based evaluation framework for prediction models, we consider individual RU trajectories present in WOMD and categorize them assigning one or more of the tags summarized in Table~\ref{tab:tags}. These tags are chosen to explore differences in performance between trajectories of different shapes (T1 and T2), different behaviors (T3-T5), different availability of observations (T6-T9), and the same or different trajectories as specified in the WMPC (T10 and T11). Future iterations of this work will include more complex scenarios (e.g. pedestrian at non-designated crossing ahead of the AV). Examples of trajectories from some of the selected tags are shown in Fig.~\ref{fig:trajectories}.
\begin{table}[hb]
\centering
\caption{Tags used to label trajectories}
\vspace{-6pt}
\label{tab:tags}
\begin{tabular}{cp{7.3cm}}
\toprule
Tag & Description \\ \midrule
T1 & \textit{Straight} - RU closely follows a straight path \\
T2 & \textit{Non-straight} - RU deviates from a straight path \\
T3 & \textit{Starting} - RU is still during observation and moves in the future \\
T4 & \textit{Stopping} - RU moves during observation and stops in the future \\
T5 & \textit{Still} - RU is still during observation and in the future \\
T6 & \textit{Late} - RU is detected late ($\leq$ 0.3 sec before prediction) \\
T7 & \textit{Very Late} - RU is detected very late (0.1 sec before prediction) \\
T8 & \textit{Full} - RU is detected during the entire observation period \\
T9 & \textit{Reappearance} - The same RU disappears during observation and reappears in the future \\
T10 & \textit{TTP} - Trajectories To Predict - Required trajectories to predict for Waymo's Motion Prediction Challenge \\
T11 & \textit{NTTP} - Trajectories that were not required for Waymo's Motion Prediction Challenge \\ \bottomrule
\end{tabular}
\end{table}

To illustrate the importance of a thorough assessment, the performance of three different models is compared, first in a manner that adheres to common current evaluation practices, and then considering additional aspects and revealing observations that are crucial for understanding the limitations of each of the methods but that could not be concluded from the initial evaluation.

\subsection{Evaluated Models}
Three different prediction models are compared, in increasing level of complexity. The inputs and outputs of each model are summarized in Table~\ref{tab:models}.

\textit{a)} A Constant Velocity (CV) model, which is often used as a baseline~\cite{Pellegrini2009YoullTracking}.
    CV assumes every RU maintains the same velocity over all prediction horizons, considering only the last observed RU state and disregarding the static (e.g. walls) and dynamic (e.g. other RUs) environment.

\textit{b)} An LSTM encoder-decoder neural network. Virtually any state of the art DL method has some recurrent neural network component to encode the past of RUs, with LSTM-based architectures being a popular choice~\cite{Varadarajan2021MultiPath++:Prediction,MunozSanchez2020APrediction}. The implemented LSTM considers past observations of RUs and does not exploit any knowledge about road topology or surrounding RUs, meaning that its predictions will not avoid overlap with other RUs or static obstacles similarly to CV.

\textit{c)} MotionCNN~\cite{Konev2021MotionCNN:Driving}, the 3rd place solution of the WMPC. MotionCNN provides an elegant solution to exploit different types of information (i.e. past of RUs, surrounding RUs, and road infrastructure) with an architecture based on convolutional neural networks only, and it produces 6 feasible trajectories. We downloaded a pre-trained model from an open repository provided by the authors\footnote{\href{https://github.com/kbrodt/waymo-motion-prediction-2021/releases}{https://github.com/kbrodt/waymo-motion-prediction-2021/releases}}.

\begin{table}[htb]
\centering
\caption{Overview of compared models and their inputs and outputs}
\vspace{-10pt}
\label{tab:models}
\resizebox{\linewidth}{!}{%
\begin{tabular}{@{}c|ccc|cc@{}}
\toprule
\multirow{3}{*}{\textbf{Model}} & \multicolumn{3}{c|}{\textbf{Input}} & \multicolumn{2}{c}{\textbf{Output}} \\
 &  Past & \multirow{2}{*}{Other RUs} & \multirow{2}{*}{Map} & Single & Multiple \\
 &  RU state &  &  & Trajectory & Trajectories \\ \midrule
CV &  &  &  & \checkmark \\
LSTM & \checkmark &  &  & \checkmark \\
MotionCNN & \checkmark & \checkmark & \checkmark & & \checkmark \\ \bottomrule
\end{tabular}%
}
\end{table}
\vspace{-12pt}

\subsection{Evaluation Criteria}
To assess the accuracy of the predictions at various prediction horizons, the horizons on which the models are evaluated start with the first future timestep (correspoding to 0.1 seconds into the future), and follow the same sampling rate of 2Hz as in the WMPC. Thus, we compute minADE and minFDE as defined in \eqref{eq:minADE} and \eqref{eq:minFDE} for $T=\{0.1 + \frac{t}{2} \mid 0 \leq t < 16 \land t \in \mathbb{Z} \}$ seconds.

% The three models are evaluated calculating the errors of their predictions at each prediction horizon. To compare models that produce different outputs (CV and LSTM predict one trajectory, while MotionCNN predicts 6 and a confidence score for each), we use minADE and minFDE.

\section{Results} \label{sec:results}
An iterative approach is taken to report the performance of the evaluated models. At each iteration, we analyze the models' performance in increasing detail and draw conclusions regarding their capabilities. 
Subsequent, more detailed iterations reveal different, possibly conflicting conclusions, showing that current evaluation methods can be misleading. 
\subsection{Overall Evaluation}
Table \ref{tab:results-overall} shows the performance of the models for trajectories of vehicles (veh.), pedestrians (ped.) and cyclists (cyc.) aggregated over all prediction horizons, for the required trajectories in the WMPC. From these results we might conclude:
\begin{myenum}
    \item[C1)] DL models are superior when compared to CV, especially for predicting vehicles' trajectories, as can be seen by the large difference in minADE and minFDE.
    \item[C2)] MotionCNN is clearly the most suitable model to predict trajectories of vehicles and cyclists.
    \item[C3)] LSTM is better suited for pedestrian trajectory prediction than the other two models.
\end{myenum}
\input{includes/tab-results-overall}
\input{includes/tab-results-overall-nttp}
\input{includes/tab-results-per-time}

Many works report the performance of their method in a similar fashion as presented in Table \ref{tab:results-overall} \cite{gu2021densetntCustom,Konev2021MotionCNN:Driving}, over all prediction horizons and only for a subset of trajectories deemed interesting, raising the question of whether or not performance remains the same for other trajectories. Table~\ref{tab:results-overall-nttp} summarizes the models' accuracies after evaluation on all trajectories not considered previously. From this table, additional conclusions can be made: 

\begin{myenum}
    \item[C4)] A simple model such as CV outperforms complex DL methods such as MotionCNN, which contradicts C1 and C2. It seems that trajectories deemed uninteresting (NTTP) originate from RUs moving at constant velocity.
    \item[C5)] MotionCNN achieves the lowest accuracy for all three types of RUs, contradicting C1 and C2. This suggests MotionCNN suffers from overfitting, since it does not generalize well to different trajectories, which is problematic since in real-world applications one does not know beforehand if a trajectory would be considered TTP or not.
    \item[C6)] LSTM presents the highest accuracy for prediction of pedestrian trajectories (C3 still holds), and additionally vehicles and possibly cyclists (contradicting C2).
    \item[C7)] The errors are significantly lower, so these trajectories are less challenging to predict. 
\end{myenum}
Following C7, the reader might wonder how it is possible that MotionCNN performs the best for selected challenging trajectories, yet it performs the worst for easier trajectories. The reason might not be obvious, since we purposely left out some practical details on how the models were trained to emphasize the importance of reporting practical development details of data-driven models. LSTM was trained using all the available trajectories, and MotionCNN using only those required to be predicted in the WMPC. As such, it is not extraordinary to observe the differences in Tables \ref{tab:results-overall} and \ref{tab:results-overall-nttp}, as evaluation on only TTP or NTTP trajectories would give an advantage to the model trained with those trajectories. 
However, this reveals yet another pitfall of standard metrics: they do not provide any indication of a model's performance for trajectories that are uncommon in the data. This generalization issue, which is quite common in data-driven models, has already been recognized in the context of motion prediction \cite{Rudenko2020HumanSurvey}, but it remains unaddressed. 
If we were to further train MotionCNN also using the unseen trajectories, its performance would improve for these, but it remains unclear if its performance on the selected trajectories would be negatively affected.

\subsection{Evaluation Per Time Horizon}
\vspace{-14pt}
A model's performance might vary significantly depending on the prediction horizon considered, so it is important take that into account, since different applications have different accuracy requirements for different horizons. Evaluating accuracy at different horizons is also common in trajectory prediction literature \cite{Deo2018ConvolutionalPrediction,Chandra2019RobustTP:Inputs}. 
Additionally, further analyzing other derived metrics such as standard deviation or maximum prediction errors, which is not commonly done, might reveal interesting insights. Table~\ref{tab:results-per-time} summarizes such an analysis (only for trajectories labeled TTP), revealing new insights that were not obvious from the previous superficial analysis from Table~\ref{tab:results-overall}:

\begin{myenum}
    \item[C8)] CV can match and even outperform complex DL methods, \textit{for very short prediction horizons}, which contradicts C1-C3 and C6, and further details C4. 
    \item[C9)] CV can no longer match the performance of the other models for horizons longer than 1 second, which adds detail to C4. 
    \item[C10)] LSTM presents the highest errors  \textit{for very short prediction horizons}, even for pedestrians, which contradicts C3 and C6.
    \item[C11)] LSTM predictions of pedestrian trajectories are the closest to the real trajectory overall, but not necessarily for long prediction horizons (as seen by the lowest minADE, but not minFDE at time 7.6 seconds).
    \item[C12)] MotionCNN is best suited for trajectory prediction of vehicles and cyclists overall, but its worst predictions can be less accurate than those of LSTM, even when considering the best out of the 6 predicted trajectories.
\end{myenum}

\noindent Several additional observations could be made from Table~\ref{tab:results-per-time}, which would either contradict or confirm the initial conclusions with a finer level of detail. Thus, it is important to always keep the intended application in mind and analyze the aspects that are most relevant for this purpose. 
\input{includes/tab-results-per-tag}

\subsection{Evaluation Per Type of Trajectory}
Next we analyze model performance for different types of trajectories according to the tags described earlier. Table~\ref{tab:results-per-tag} shows the results and new observations can be made: 
 
 \input{includes/fig-tag-counts}
 \begin{myenum}
    \item[C13)] Some behavior is missing in the trajectories used for evaluation. There are not pedestrian or cyclist trajectories where the RU stands still during the observation period and either remains still or moves in the next 8 seconds\footnote{An RU was considered still if its speed did not exceed 0.01 m/s.} (denoted by ``-'' in Table~\ref{tab:results-per-tag}). Additionally, there are no trajectories where the RU disappears shortly before making the prediction. Thus, if an AV is presented with these situations, it will not be possible to determine which model is most suitable. 
    \item[C14)] CV is most suitable for predictions where the RU remains still (trivial). MotionCNN performs the worst by far in these cases. If an AV is presented with this situation, it might use a model that is not the most appropriate in this case despite being the most accurate overall.
    \item[C15)] CV is most suitable for predicting starting behavior. This conclusion could be misleading, as CV naively predicts the RU remains still, and it does not predict starting behavior. If the RU remains still for most of the future 8 seconds, the overall error will be low, but it does not capture CV's ability to predict when movement will start.
    \item[C16)] LSTM is particularly well suited for prediction of all types of pedestrian trajectories, which supports some of our previous conclusions (i.e. C3, C6 and partially C11), but contradicts some others (i.e. C8, C10 and C14).
    \item[C17)] MotionCNN is well suited for prediction of trajectories labeled TTP, straight trajectories of vehicles and cyclists, and non-straight trajectories of vehicles. However, for other types of trajectories, such as late detections or trajectories labeled NTTP, it can even be outperformed by CV. 
\end{myenum}

After analyzing model performance considering different aspects like various prediction horizons and trajectory types, it is still not possible to conclude on the potential suitability of a data-driven model, since their performance is heavily affected by the data used to train them. Recall that MotionCNN was trained to predict only those trajectories labeled TTP, while LSTM was trained using all trajectories. Figure~\ref{fig:tag-counts} shows the percentage of trajectories labeled with each tag, both for the entire dataset and only considering trajectories labeled TTP, which partly explains the difference in performance between LSTM and MotionCNN:
\begin{myenum}
    \item [C18)] LSTM outperforms MotionCNN for most trajectory types because it has been trained with more instances of each type. MotionCNN was trained on trajectories labeled TTP, which are only about 7\% of available data.
    \item [C19)] Within the sets of trajectories used to train each model, the relative frequency of some behaviors is very different. For instance, MotionCNN is unable to predict starting and still trajectories accurately because this behavior is significantly underrepresented in TTP trajectories (approximately 0.08 and 0.1\%), as opposed to their frequency in the entire dataset used to train LSTM (18 and 53\%).
\end{myenum}

Even if a model's performance can be better explained by an in-depth analysis of the training data, it does not mean it is always necessary to do so. If a model will only be used in specific situations (e.g. late detections), then it must be accurate in these situations no matter how innacurate it might be in others. Similarly, if a model's purpose is to increase safety or fuel efficiency, then vehicle-level performance metrics after integrating this prediction model in the vehicle should be the main assessment criteria, since a marginal improvement in predictive accuracy might yield little to no improvement for the intended application. Thus, evaluation of these models should be done according to the application in which they operate.

\section{Conclusion and Future Work} \label{sec:conclusion}
To operate safely, an AV must anticipate the future motion of other RUs in its surroundings through trajectory prediction. Assessment of prediction models is commonly performed over a set of trajectories without distinction over the type of movement captured by each trajectory, which does not provide a clear overview of the suitability of each model for different situations. Furthermore, the impact of a marginal increase in predictive accuracy at the vehicle level remains unclear, as other components of an AV are normally not considered for assessment of prediction models.

% [Updated text] In this work, we have illustrated the extent to which standard evaluation practices result in misleading conclusions of a model's predictive capabilities. Additionally, we have taken a first step towards a scenario-based framework for evaluation of prediction models, classifying individual trajectories with various tags according to the type of movement they capture and allowing a clear assessment of a model's suitability to predict each trajectory type. 

In this work, we have illustrated the extent to which standard evaluation practices result in misleading conclusions of a model's predictive capabilities.
Additionally, we have made publicly available a scenario-based framework for evaluation of prediction models, which allows classification of individual trajectories according to the type of movement they capture and facilitates a clear assessment of a model's suitability to predict each trajectory type. 

% The implemented code is publicly available\footnote{Code will be published on GitHub and shared in the camera-ready version.} to encourage a thorough assessment of prediction models. 

Future work will extend the framework proposed in \cite{DeGelder2022TowardsFrameworkCust}
to model and capture relevant scenarios for prediction algorithms, facilitating assessment for their intended application.

% \balance

% \addtolength{\textheight}{-15cm}   % This command serves to balance the column lengths

%%%%%%%%%%%%%%%%%%%%%%%%%%%%%%%%%%%%%%%%%%%%%%%%%%%%%%%%%%%%%%%%%%%%%%%%%%%%%%%%

%%%%%%%%%%%%%%%%%%%%%%%%%%%%%%%%%%%%%%%%%%%%%%%%%%%%%%%%%%%%%%%%%%%%%%%%%%%%%%%%

% \textcolor{red}{[If space allows, add more details on our LSTM and the version of MotionCNN used. If space does not allow, it would be interesting to make a explicit reference stating this is the section where we would provide more details, but we will not in order to avoid the very expensive extra page charges, which is probably part of the reason practical details are often lacking in publications (seems like a terrible idea to get the paper published, so I probably won't do it).]}

% \section*{ACKNOWLEDGMENT}

% The preferred spelling of the word ``acknowledgment'' in America is without an ``e'' after the ``g''. Avoid the stilted expression, ``One of us (R. B. G.) thanks . . .''  Instead, try ``R. B. G. thanks''. Put sponsor acknowledgments in the unnumbered footnote on the first page.

%%%%%%%%%%%%%%%%%%%%%%%%%%%%%%%%%%%%%%%%%%%%%%%%%%%%%%%%%%%%%%%%%%%%%%%%%%%%%%%%

\printbibliography

% \bibliographystyle{IEEEtran}
% \bibliography{IEEEabrv,references_custom,references}

\end{document}

%% file: includes/tab-results-overall.tex
\begin{table}[b]
\vspace{-12pt}
\centering
\caption{Model performance over all prediction horizons (TTP)}
\vspace{-10pt}
\label{tab:results-overall}
\resizebox{\linewidth}{!}{%
\begin{tabular}{@{}c|lll|lll@{}}
\toprule
\multirow{3}{*}{\textbf{Model}} & \multicolumn{6}{c}{\textbf{Metric}} \\
 & \multicolumn{3}{c|}{minADE} & \multicolumn{3}{c}{minFDE} \\
 & \multicolumn{1}{c}{Veh.} & \multicolumn{1}{c}{Ped.} & \multicolumn{1}{c|}{Cyc.} & \multicolumn{1}{c}{Veh.} & \multicolumn{1}{c}{Ped.} & \multicolumn{1}{c}{Cyc.} \\ \midrule
 
CV & \textcolor{red}{3.893} & \textcolor{red}{0.633} & \multicolumn{1}{l|}{\textcolor{red}{1.632}} & \multicolumn{1}{c}{\textcolor{red}{10.468}} & \textcolor{red}{1.477} & \textcolor{red}{3.964} \\
LSTM & 2.025 & \textcolor{blue}{0.545} & \multicolumn{1}{l|}{1.481} & 5.681 & \textcolor{blue}{1.295} & 3.618 \\ 
MotionCNN & \textcolor{blue}{1.482} & 0.564 & \multicolumn{1}{l|}{\textcolor{blue}{1.247}} & \textcolor{blue}{3.855} & 1.315 & \textcolor{blue}{3.012} \\ \bottomrule
\multicolumn{7}{c}{\textcolor{red}{Red} and \textcolor{blue}{blue} indicate highest and lowest error per RU}
\end{tabular}%
}
% \vspace{-10pt}
\end{table}

%% file: includes/tab-results-overall-nttp.tex
\begin{table}[htb]
\centering
\renewcommand\thetable{V}
\caption{Model performance over all prediction horizons (NTTP)}
\vspace{-10pt}
\label{tab:results-overall-nttp}
\resizebox{\linewidth}{!}{%
\begin{tabular}{@{}c|lll|lll@{}}
\toprule
\multirow{3}{*}{\textbf{Model}} & \multicolumn{6}{c}{\textbf{Metric}} \\
 & \multicolumn{3}{c|}{minADE} & \multicolumn{3}{c}{minFDE} \\
 & \multicolumn{1}{c}{Veh.} & \multicolumn{1}{c}{Ped.} & \multicolumn{1}{c|}{Cyc.} & \multicolumn{1}{c}{Veh.} & \multicolumn{1}{c}{Ped.} & \multicolumn{1}{c}{Cyc.} \\ \midrule
 
CV & 0.640
 & 0.565
 & 1.070
 & 1.834
 & 1.284
 & 2.816
 \\

LSTM & \textcolor{blue}{0.391} 
& \textcolor{blue}{0.320} 
& \textcolor{blue}{1.014}
& \textcolor{blue}{1.114}
 & \textcolor{blue}{0.728} 
 &  \textcolor{blue}{2.441}
\\

MotionCNN & \textcolor{red}{0.806} 
& \textcolor{red}{0.637} 
& \textcolor{red}{1.178}
& \textcolor{red}{2.297} 
& \textcolor{red}{1.575} 
&\textcolor{red}{3.291} \\ \bottomrule
\end{tabular}%
}
% \vspace{-18pt}
\end{table}

%% file: includes/tab-results-per-time.tex
\begin{table*}[hb]
% \vspace{-10pt}
\renewcommand\thetable{VI}
\caption{Model Performance At Different Prediction Horizons (Only Trajectories Labeled TTP)}
\vspace{-10pt}
\label{tab:results-per-time}
\resizebox{\linewidth}{!}{%
\begin{tabular}{ccc|ccc|ccc|ccc|ccc|ccc}
\toprule
 &  & Time [s] & \multicolumn{3}{c}{0.1} & \multicolumn{3}{c}{1.1} & \multicolumn{3}{c}{3.1} & \multicolumn{3}{c}{5.1} & \multicolumn{3}{c}{7.6} \\
 &  &  &  & std  & max &  & std  & max &  & std  & max &  & std  & max &  & std  & max \\
Model & RU & Metric [m] &  &   &  &  &  &  &  &  &  &  &  &  &  &  &  \\ \midrule
\multirow[c]{6}{*}{CV} & \multirow[c]{2}{*}{Veh.} & minADE & \color{blue} 0.034 & 0.084 & 6.90 & \color{red} 0.393 & \color{red} 0.717 & \color{red} 72.889 & \color{red} 2.187 & \color{red} 3.021 & \color{red} 205.598 & \color{red} 5.26 & \color{red} 6.805 & \color{red} 336.425 & \color{red} 9.878 & \color{red} 12.312 & \color{red} 465.797 \\
 &  & minFDE & \color{blue} 0.034 & 0.084 & 6.90 & \color{red} 0.842 & \color{red} 0.971 & \color{red} 72.889 & \color{red} 5.663 & \color{red} 4.303 & \color{red} 205.598 & \color{red} 14.142 & \color{red} 9.796 & \color{red} 336.425 & \color{red} 27.323 & \color{red} 17.929 & \color{red} 465.797 \\ \cmidrule{2-18}
 & \multirow[c]{2}{*}{Ped.} & minADE & 0.023 & 0.025 & 0.825 & \color{red} 0.124 & \color{red} 0.163 & \color{red} 6.123 & \color{red} 0.434 & \color{red} 0.573 & \color{red} 17.715 & \color{red} 0.847 & \color{red} 1.114 & \color{red} 29.263 & \color{red} 1.415 & \color{red} 1.858 & \color{red} 43.429 \\
 &  & minFDE & 0.023 & 0.025 & 0.825 & \color{red} 0.235 & \color{red} 0.211 & \color{red} 6.123 & \color{red} 0.969 & \color{red} 0.85 & \color{red} 17.715 & \color{red} 1.974 & \color{red} 1.703 & \color{red} 29.263 & \color{red} 3.473 & \color{red} 2.929 & \color{red} 43.429 \\ \cmidrule{2-18}
 & \multirow[c]{2}{*}{Cyc.} & minADE & 0.054 & 0.048 & 0.449 & 0.285 & \color{red} 0.333 & \color{red} 3.567 & \color{red} 1.061 & \color{red} 1.59 & \color{blue} 53.026 & \color{red} 2.177 & \color{red} 3.017 & \color{blue} 58.425 & \color{red} 3.798 & \color{red} 5.186 & \color{blue} 67.002 \\
 &  & minFDE & 0.054 & 0.048 & 0.449 & \color{red} 0.539 & \color{red} 0.41 & \color{red} 3.567 & \color{red} 2.431 & \color{red} 2.278 & \color{blue} 51.616 & \color{red} 5.271 & \color{red} 4.623 & \color{blue} 58.425 & \color{red} 9.659 & \color{red} 8.238 & \color{blue} 67.002 \\ \midrule
\multirow[c]{6}{*}{LSTM} & \multirow[c]{2}{*}{Veh.} & minADE & \color{red} 0.051 & \color{red} 0.109 & \color{red} 9.492 & 0.223 & \color{blue} 0.425 & \color{blue} 66.493 & 1.076 & 1.617 & \color{blue} 68.542 & 2.655 & 3.952 & \color{blue} 68.542 & 5.313 & 7.957 & \color{blue} 119.801 \\
 &  & minFDE & \color{red} 0.051 & \color{red} 0.109 & \color{red} 9.492 & 0.435 & \color{blue} 0.564 & \color{blue} 65.049 & 2.762 & 2.481 & \color{blue} 61.582 & 7.368 & 6.487 & \color{blue} 63.179 & 15.828 & 13.615 & \color{blue} 119.801 \\ \cmidrule{2-18}
 & \multirow[c]{2}{*}{Ped.} & minADE & \color{red} 0.059 & \color{red} 0.056 & \color{red} 2.79 & 0.114 & \color{blue} 0.127 & 2.852 & \color{blue} 0.36 & \color{blue} 0.514 & 8.716 & \color{blue} 0.72 & \color{blue} 1.04 & \color{blue} 14.252 & \color{blue} 1.236 & 1.761 & \color{blue} 25.532 \\
 &  & minFDE & \color{red} 0.059 & \color{red} 0.056 & \color{red} 2.79 & \color{blue} 0.183 & \color{blue} 0.173 & 2.541 & \color{blue} 0.808 & 0.812 & 8.716 & \color{blue} 1.719 & 1.656 & \color{blue} 14.252 & 3.124 & 2.859 & \color{blue} 25.532 \\ \cmidrule{2-18}
 & \multirow[c]{2}{*}{Cyc.} & minADE & \color{red} 0.216 & \color{red} 0.184 & \color{red} 2.558 & \color{red} 0.30 & 0.272 & 2.925 & 0.923 & 1.496 & \color{red} 54.011 & 1.936 & 2.887 & \color{red} 62.589 & 3.462 & 5.004 & \color{red} 72.982 \\
 &  & minFDE & \color{red} 0.216 & \color{red} 0.184 & \color{red} 2.558 & 0.428 & 0.347 & 2.925 & 2.109 & 2.23 & \color{red} 54.011 & 4.793 & 4.555 & \color{red} 62.589 & 9.029 & 8.047 & \color{red} 72.982 \\ \midrule
\multirow[c]{6}{*}{MotionCNN} & \multirow[c]{2}{*}{Veh.} & minADE & 0.038 & \color{blue} 0.058 & \color{blue} 4.487 & \color{blue} 0.219 & 0.432 & 66.633 & \color{blue} 0.888 & \color{blue} 1.404 & 68.55 & \color{blue} 1.973 & \color{blue} 3.138 & 94.614 & \color{blue} 3.64 & \color{blue} 6.001 & 150.044 \\
 &  & minFDE & 0.038 & \color{blue} 0.058 & \color{blue} 4.487 & \color{blue} 0.419 & 0.584 & 65.263 & \color{blue} 2.141 & \color{blue} 2.198 & 63.885 & \color{blue} 5.089 & \color{blue} 5.358 & 94.614 & \color{blue} 10.08 & \color{blue} 11.026 & 150.044 \\ \cmidrule{2-18}
 & \multirow[c]{2}{*}{Ped.} & minADE & \color{blue} 0.022 & \color{blue} 0.019 & \color{blue} 0.477 & \color{blue} 0.108 & 0.134 & \color{blue} 1.723 & 0.384 & 0.523 & \color{blue} 7.25 & 0.758 & 1.049 & 16.647 & 1.261 & \color{blue} 1.745 & 35.719 \\
 &  & minFDE & \color{blue} 0.022 & \color{blue} 0.019 & \color{blue} 0.477 & 0.202 & \color{blue} 0.173 & \color{blue} 1.723 & 0.868 & \color{blue} 0.801 & \color{blue} 7.25 & 1.777 & \color{blue} 1.654 & 16.647 & \color{blue} 3.05 & \color{blue} 2.822 & 35.719 \\ \cmidrule{2-18}
 & \multirow[c]{2}{*}{Cyc.} & minADE & \color{blue} 0.047 & \color{blue} 0.039 & \color{blue} 0.321 & \color{blue} 0.215 & \color{blue} 0.257 & \color{blue} 2.803 & \color{blue} 0.814 & \color{blue} 1.429 & 53.809 & \color{blue} 1.67 & \color{blue} 2.676 & 62.243 & \color{blue} 2.887 & \color{blue} 4.578 & 72.027 \\
 &  & minFDE & \color{blue} 0.047 & \color{blue} 0.039 & \color{blue} 0.321 & \color{blue} 0.404 & \color{blue} 0.326 & \color{blue} 2.803 & \color{blue} 1.874 & \color{blue} 2.12 & 53.809 & \color{blue} 4.062 & \color{blue} 4.365 & 62.243 & \color{blue} 7.213 & \color{blue} 7.775 & 72.027 \\
 \bottomrule
\end{tabular}
}
\end{table*}

%% file: includes/tab-results-per-tag.tex
\begin{table*}[b]
\vspace{-10pt}
\centering
\caption{Model Performance For Different Types of Trajectories}
\vspace{-10pt}
\label{tab:results-per-tag}
\resizebox{\linewidth}{!}{%
\begin{tabular}{ccc|ccccccccccc}
\toprule
 &  &  & T1 & T2 & T3 & T4 & T5 & T6 & T7 & T8 & T9 & T10 & T11 \\
Model & RU & Metric [m] & \scriptsize Straight & \scriptsize NonStraight & \scriptsize Starting & \scriptsize Stopping & \scriptsize Still &  \scriptsize Late& \scriptsize VeryLate & \scriptsize Full & \scriptsize Reappear & \scriptsize TTP & \scriptsize NTTP \\ \midrule
\multirow[c]{6}{*}{CV} & \multirow[c]{2}{*}{Veh.} & minADE & \color{red} 2.872 & \color{red} 2.747 & \color{blue} 0.821 & \color{red} 1.442 & \color{blue} 0 & 1.387 & \color{blue} 1.316 & \color{red} 1.089 & - & \color{red} 3.893 & 0.64 \\
 &  & minFDE & \color{red} 8.422 & \color{red} 7.285 & \color{blue} 2.847 & \color{red} 3.73 & \color{blue} 0 & 4.395 & 4.175 & \color{red} 3.15 & - & \color{red} 10.468 & 1.834 \\ \cmidrule{2-14}
 & 
 \multirow[c]{2}{*}{Ped.} & minADE & 0.466 & 0.974 & - & 0.43 & - & 1.092 & 1.102 & 0.486 & - & \color{red} 0.633 & 0.565 \\ 
 &  & minFDE & \color{red} 1.127 & 2.226 & - & 0.908 & - & 2.721 & 2.752 & 1.152 & - & \color{red} 1.477 & 1.284 \\ \cmidrule{2-14}
 & 
 \multirow[c]{2}{*}{Cyc.} & minADE & \color{red} 1.445 & \color{red} 2.004 & - & 1.071 & - & \color{blue} 1.91 & \color{blue} 2.037 & \color{red} 1.381 & - & \color{red} 1.632 & 1.07 \\
 &  & minFDE & \color{red} 3.67 & 5.208 & - & 2.357 & - & 4.976 & 5.38 & \color{red} 3.54 & - & \color{red} 3.964 & 2.816 \\ \midrule 
\multirow[c]{6}{*}{LSTM} & \multirow[c]{2}{*}{Veh.} & minADE & 1.549 & 1.693 & 0.828 & \color{blue} 0.581 & 0.03 & \color{blue} 1.364 & 1.449 & \color{blue} 0.574 & - & 2.025 & \color{blue} 0.391 \\
 &  & minFDE & 4.679 & 4.748 & 2.853 & \color{blue} 1.608 & 0.027 & \color{blue} 3.692 & \color{blue} 3.642 & \color{blue} 1.739 & - & 5.681 & \color{blue} 1.114 \\ \cmidrule{2-14}
 & \multirow[c]{2}{*}{Ped.} & minADE & \color{blue} 0.401 & \color{blue} 0.597 & - & \color{blue} 0.182 & - & \color{blue} 0.715 & \color{blue} 0.843 & \color{blue} 0.344 & - & \color{blue} 0.545 & \color{blue} 0.32 \\
 &  & minFDE & \color{blue} 0.973 & \color{blue} 1.495 & - & \color{blue} 0.38 & - & \color{blue} 1.619 & \color{blue} 1.835 & \color{blue} 0.845 & - & \color{blue} 1.295 & \color{blue} 0.728 \\ \cmidrule{2-14}
 & \multirow[c]{2}{*}{Cyc.} & minADE & 1.327 & \color{blue} 1.93 & - & \color{blue} 0.967 & - & 2.308 & \color{red} 3.143 & 1.234 & - & 1.481 & \color{blue} 1.014 \\
 &  & minFDE & 3.304 & \color{blue} 5.027 & - & \color{blue} 2.142 & - & \color{blue} 4.39 & \color{blue} 5.129 & 3.203 & - & 3.618 & \color{blue} 2.441 \\ \midrule
\multirow[c]{6}{*}{MotionCNN} & \multirow[c]{2}{*}{Veh.} & minADE & \color{blue} 1.342 & \color{blue} 1.297 & \color{red} 1.058 & 0.861 & \color{red} 0.629 & \color{red} 1.868 & \color{red} 2.745 & 0.836 & - & \color{blue} 1.482 & \color{red} 0.806 \\
 &  & minFDE & \color{blue} 3.633 & \color{blue} 3.331 & \color{red} 3.109 & 2.40 & \color{red} 1.861 & \color{red} 4.483 & \color{red} 6.232 & 2.429 & - & \color{blue} 3.855 & \color{red} 2.297 \\ \cmidrule{2-14}
 & \multirow[c]{2}{*}{Ped.} & minADE & \color{red} 0.478 & \color{red} 0.981 & - & \color{red} 0.764 & - & \color{red} 1.748 & \color{red} 2.372 & \color{red} 0.506 & - & 0.564 & \color{red} 0.637 \\
 &  & minFDE & 1.126 & \color{red} 2.425 & - & \color{red} 1.833 & - & \color{red} 4.285 & \color{red} 5.708 & \color{red} 1.277 & - & 1.315 & \color{red} 1.575 \\ \cmidrule{2-14}
 & \multirow[c]{2}{*}{Cyc.} & minADE & \color{blue} 1.19 & 1.994 & - & \color{red} 1.261 & - & \color{red} 2.408 & 2.399 & \color{blue} 1.072 & - & \color{blue} 1.247 & \color{red} 1.178 \\
 &  & minFDE & \color{blue} 2.854 & \color{red} 5.222 & - & \color{red} 3.185 & - & \color{red} 5.967 & \color{red} 6.104 & \color{blue} 2.782 & - & \color{blue} 3.012 & \color{red} 3.291 \\
 \bottomrule
% \multicolumn{14}{c}{``-'' Indicates there were no trajectories of this type for a given RU}
\end{tabular}}
\end{table*}

%% file: includes/fig-tag-counts.tex
\begin{figure*}[htb]
% \vspace{10pt}
    \centering
    \includegraphics[width=\linewidth]{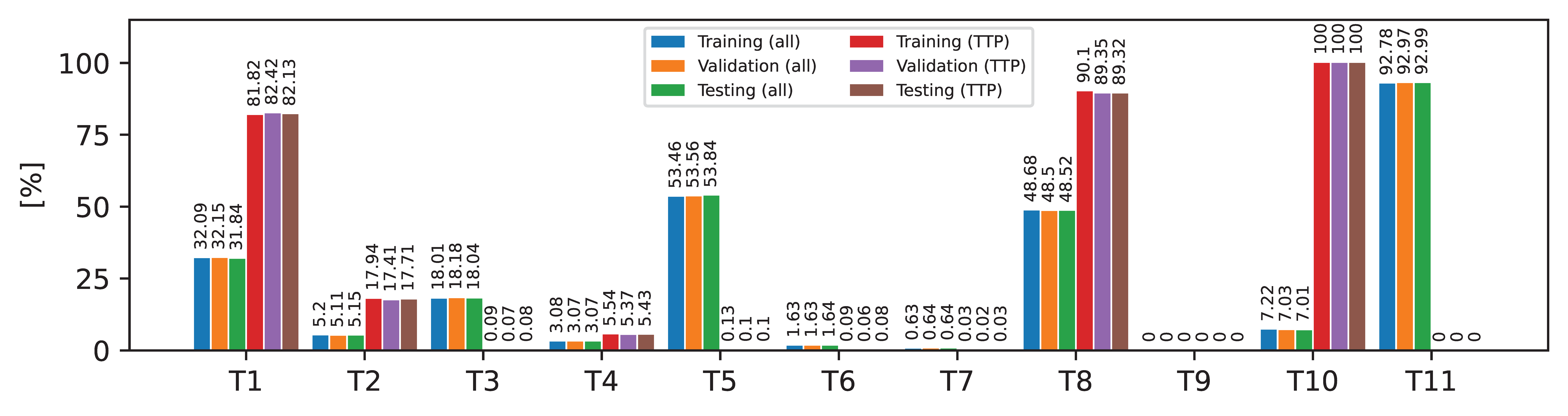}
    % \vspace{-28pt}
    \caption{Percentage of trajectories with a specific tag for the entire dataset and the set of trajectories labeled TTP.}
    \label{fig:tag-counts}
    % \vspace{-10pt}
    % \vspace{20pt}
\end{figure*}